\documentclass[letterpaper, 10 pt, conference]{ieeeconf}

\IEEEoverridecommandlockouts

\usepackage{graphicx}
\usepackage{amsmath}
\usepackage{amssymb}
\usepackage{algorithm}
\usepackage{algorithmic}
\usepackage{subfig}
\usepackage{xcolor}
\usepackage{graphicx}
\usepackage{array}
\usepackage{tikz}

\usepackage{multirow}
\usepackage{multicol}
\usepackage{booktabs}
\usepackage{diagbox}
\usepackage{makecell}
\usepackage{subcaption}
\usepackage{latexml}

\usetikzlibrary{positioning,fit,arrows.meta,backgrounds,calc}

\title{\LARGE \bf
Reusing Locomotion Training Recipes Across Humanoids \\ via Body-Aligned Virtual Joints
}

\iflatexml
\makeatletter
\@ifundefined{lx@add@author}{%
\author{Sol Choi \textasteriskcentered\\Korea Institute of Science and Technology (KIST)\\Yonsei University
\and Mingon Kim \textasteriskcentered\\LG Electronics
\and Seokin Seo\\Korea Institute of Science and Technology (KIST)
\and Jongeun Choi\\Yonsei University
\and Donghyeon Kim \dag\\1X Technologies\\kdh0429@snu.ac.kr
\and Yisoo Lee \dag\\Korea Institute of Science and Technology (KIST)\\yisoo.lee@kist.re.kr}%
}{%
\lx@add@author{Sol Choi\textsuperscript{*}}%
\lx@add@affiliation{Korea Institute of Science and Technology (KIST)}%
\lx@add@affiliation{Yonsei University}%
\lx@add@author{Mingon Kim\textsuperscript{*}}%
\lx@add@affiliation{LG Electronics}%
\lx@add@author{Seokin Seo}%
\lx@add@affiliation{Korea Institute of Science and Technology (KIST)}%
\lx@add@author{Jongeun Choi}%
\lx@add@affiliation{Yonsei University}%
\lx@add@author{Donghyeon Kim\textsuperscript{\dag}}%
\lx@add@affiliation{1X Technologies}%
\lx@add@email{kdh0429@snu.ac.kr}%
\lx@add@author{Yisoo Lee\textsuperscript{\dag}}%
\lx@add@affiliation{Korea Institute of Science and Technology (KIST)}%
\lx@add@email{yisoo.lee@kist.re.kr}%
}
\makeatother
\newcommand{\htmlauthornotes}{{\small\noindent\textsuperscript{*}Equal contribution. \textsuperscript{\dag}Corresponding authors.}\par}
\else
\author{Sol Choi, Mingon Kim, Seokin Seo, Jongeun Choi, Donghyeon Kim, and Yisoo Lee%
\thanks{S. Choi, S. Seo, and Y. Lee are with the Center for Humanoid Research, Korea Institute of Science and Technology, 02792, Seoul, South Korea
}%
\thanks{M. Kim is with the Advanced Robotics Lab., CTO Division, LG Electronics Inc., 06772 Seoul, South Korea
}%
\thanks{S. Choi and J. Choi are with the Department of Mechanical Engineering, Yonsei University, 03722, Seoul, South Korea
}%
\thanks{D. Kim is with the 1X Technologies, 150 Industrial Rd, San Carlos, CA 94070, USA
}%
\thanks{S. Choi and M. Kim contributed equally as co-first authors.
}%
\thanks{Yisoo Lee and Donghyeon Kim contributed equally as co-corresponding authors  {\tt\small yisoo.lee@kist.re.kr}, {\tt\small kdh0429@snu.ac.kr}.}
}
\newcommand{\htmlauthornotes}{}
\fi

\begin{document}
\maketitle
\thispagestyle{empty}
\pagestyle{empty}
\htmlauthornotes

\begin{abstract}
Reinforcement learning has enabled robust humanoid locomotion, yet training
recipes---rewards, architectures, hyperparameters---remain hard to reuse
across robots with different native joint coordinates, since policies
output native joint angles while action scales, bounds, gains, and
penalties are defined in those robot-specific coordinates. Existing
transmission models convert native targets into actuator commands but do
not standardize the coordinate interface exposed to the policy. We
introduce a \emph{body-aligned virtual joint space} upstream of this
mapping: policy actions and observations follow a common anatomical
convention, and a deterministic kinematic layer maps virtual targets to
native ones, leaving the existing transmission and low-level control stack
unchanged. Under a single unmodified recipe on three humanoids with
different leg-axis layouts, it is the only interface that tracks
accurately, keeps the feet aligned with the travel direction, and trains
stably on all three, and on Unitree G1 hardware it reduces heading drift
by nearly a factor of six.
\end{abstract}

\section{Introduction}
\label{sec:introduction}

Reinforcement learning (RL) has enabled legged robots to acquire dynamic behaviors that transfer from simulation to hardware. Learned controllers have demonstrated agile quadrupedal locomotion~\cite{tan2018simtoreal,hwangbo2019agile}, traversal over challenging terrain~\cite{lee2020terrain,miki2022perceptive}, rapid adaptation to changing dynamics~\cite{kumar2021rma}, and training within minutes using massively parallel simulation~\cite{rudin2022minutes}. Similar techniques have produced dynamic bipedal walking~\cite{xie2020cassie}, stair traversal~\cite{siekmann2021stairs}, and real-world humanoid locomotion~\cite{radosavovic2024humanoid,gu2024humanoidgym}. Nevertheless, adapting a controller to a robot with a different joint layout often requires substantial retraining and retuning by different training recipes.

\begin{figure}[t]
\centering
\setlength{\tabcolsep}{1pt}
\begin{tabular}{@{} >{\centering\arraybackslash}m{0.6cm}
                     @{\hspace{1pt}} m{0.31\columnwidth}
                     @{\hspace{1pt}} m{0.31\columnwidth}
                     @{\hspace{1pt}} m{0.31\columnwidth} @{}}
  & \centering Unitree G1 & \centering PND Adam & \centering Custom-built Platform \tabularnewline[4pt]
\rotatebox{90}{\small Native Coord.} &
  \includegraphics[width=\linewidth]{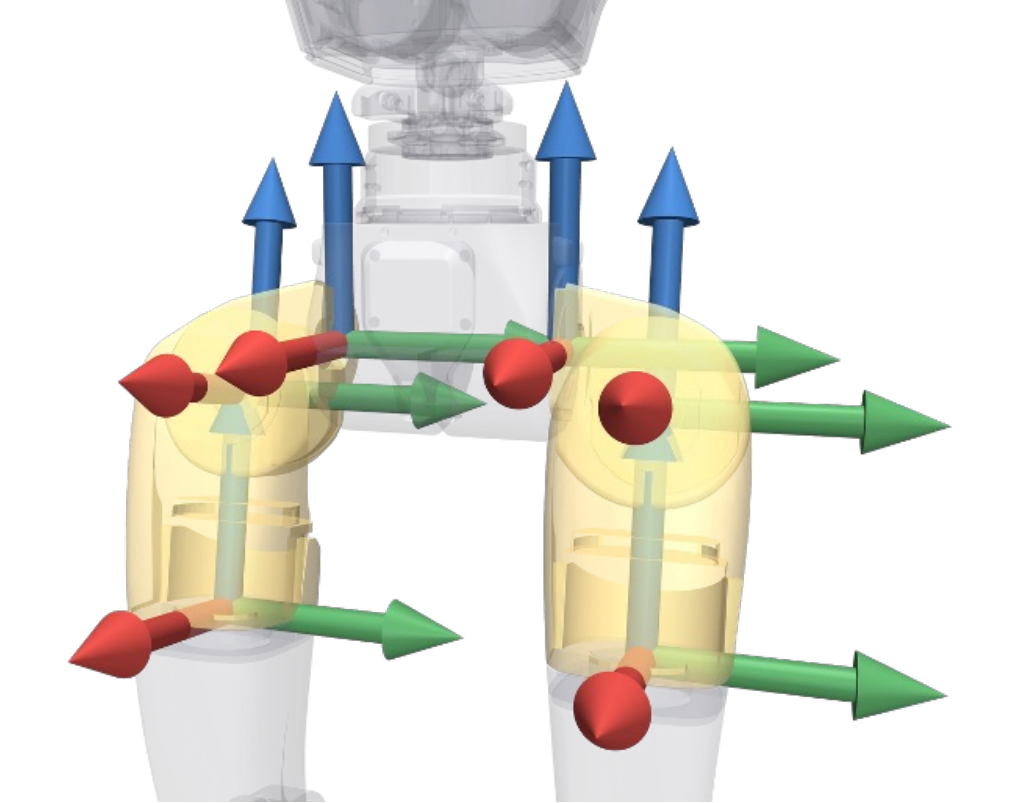}   &
  \includegraphics[width=\linewidth]{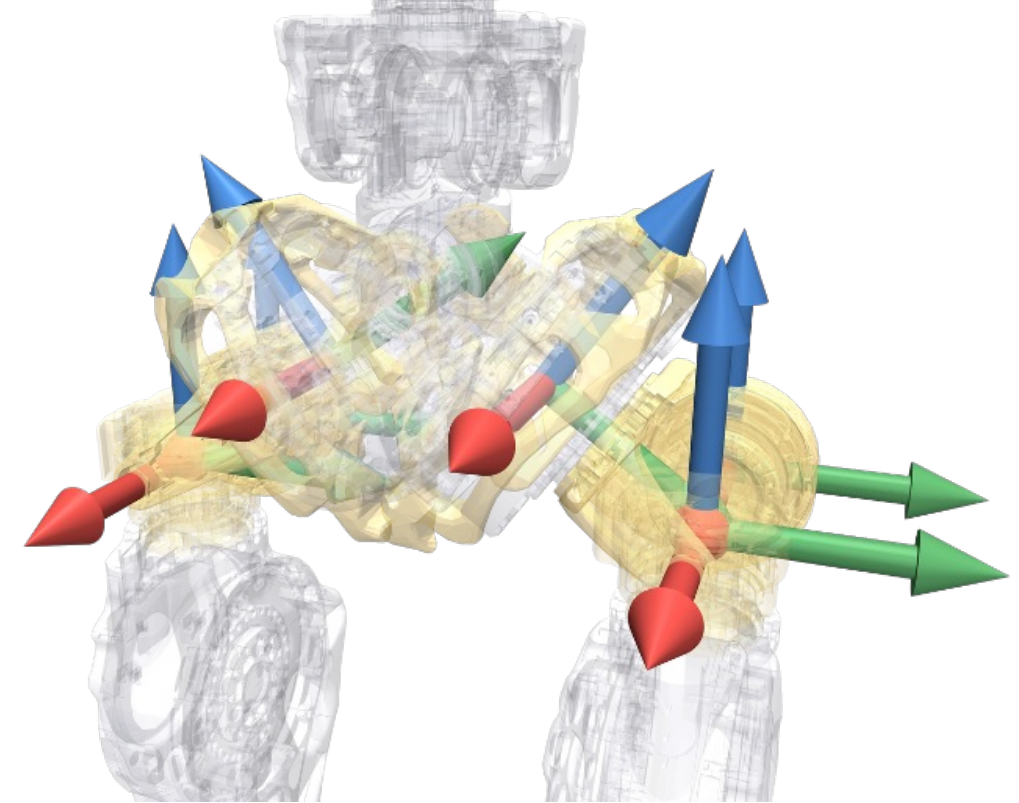}   &
  \includegraphics[width=\linewidth]{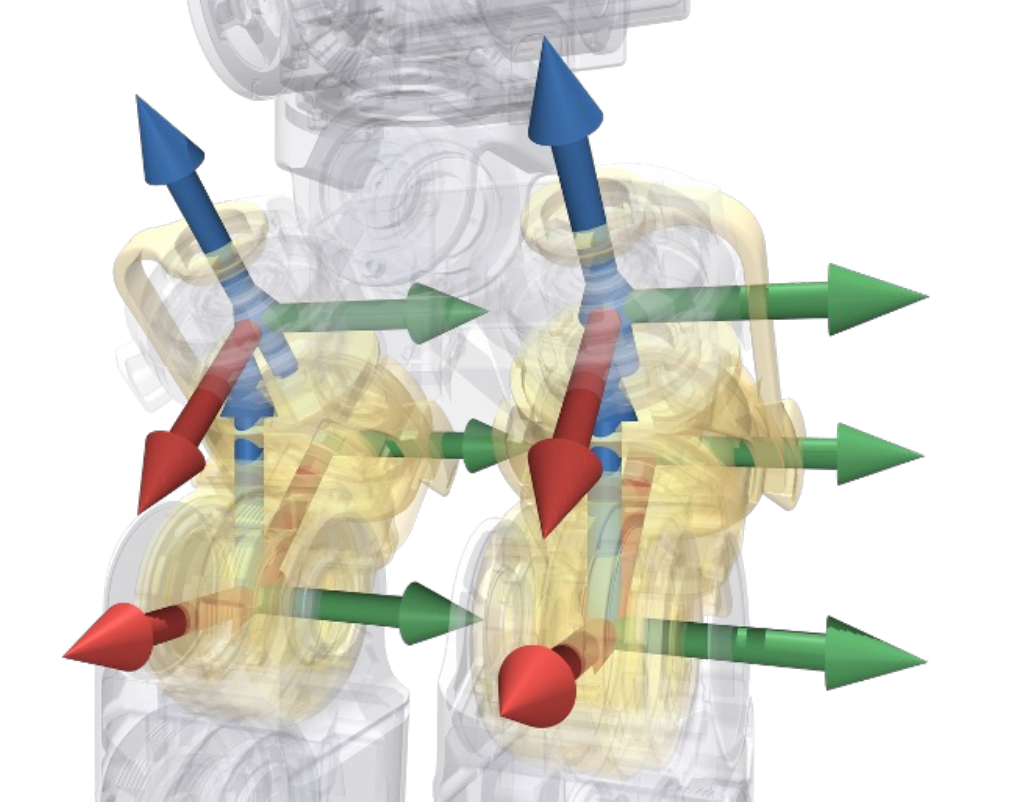}\\[2pt]
\rotatebox{90}{\small Proposed Coord.} &
  \includegraphics[width=\linewidth]{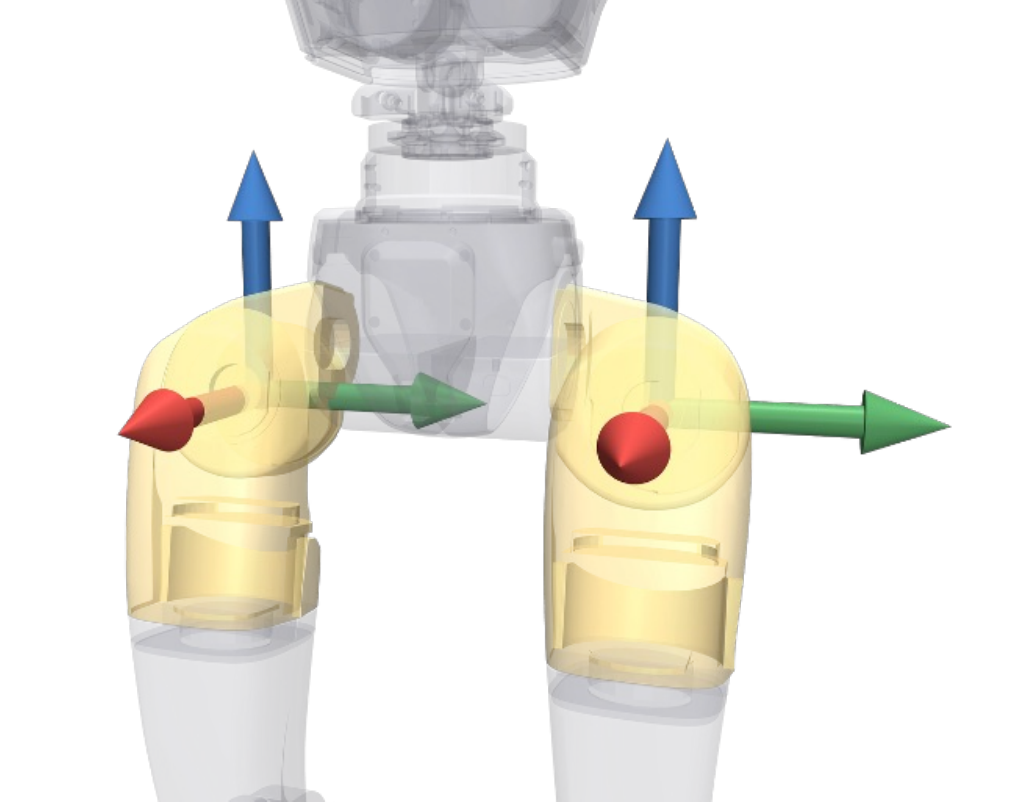}   &
  \includegraphics[width=\linewidth]{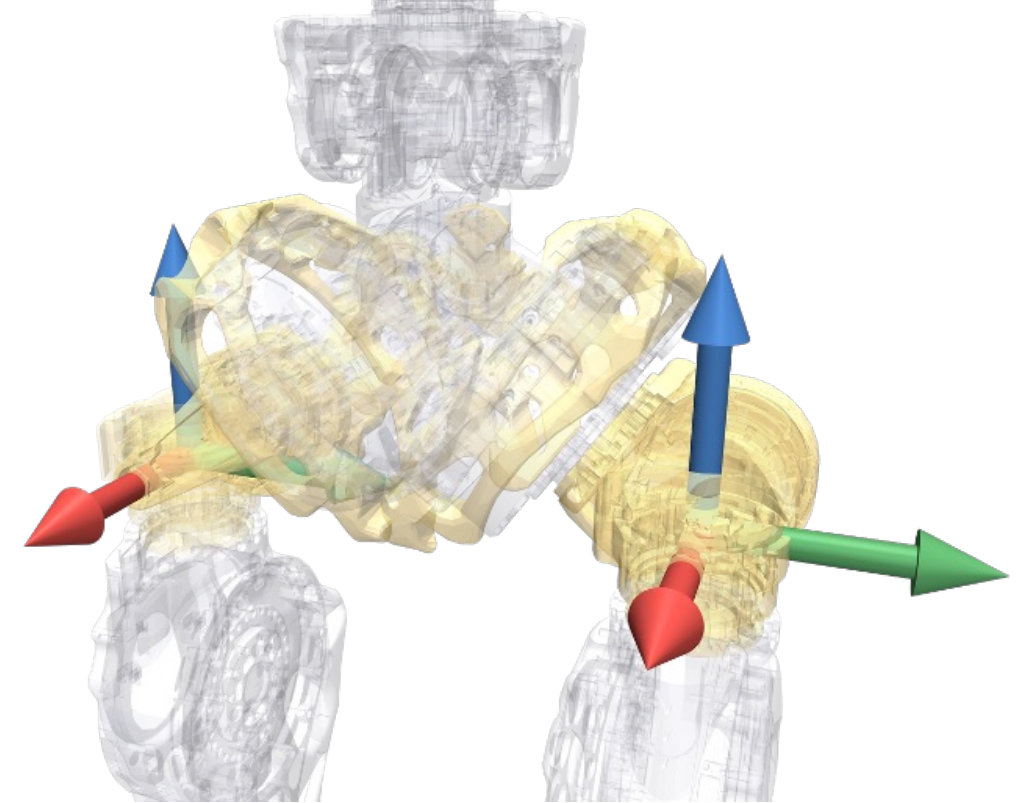}   &
  \includegraphics[width=\linewidth]{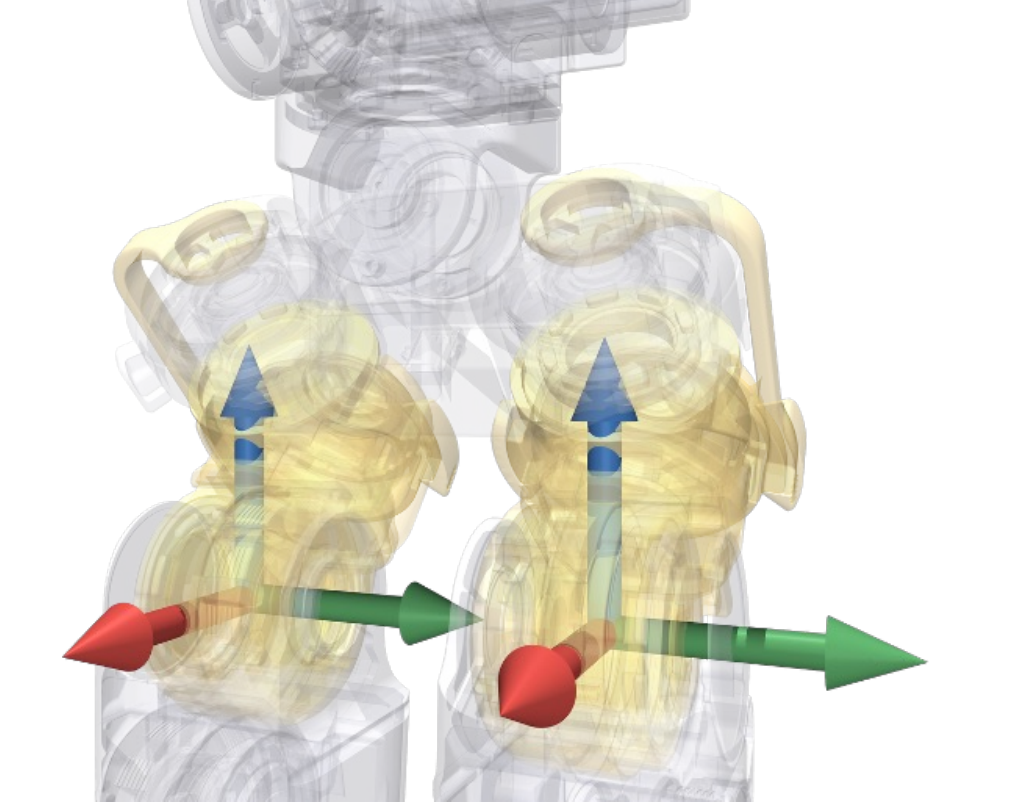}\\
\end{tabular}
\caption{Hip-joint coordinates of three representative humanoids. The native axes differ across platforms (top), whereas the proposed body-aligned virtual joint coordinates express the same joints about a shared body convention, giving a consistent axis layout across embodiments (bottom).
}
\label{fig:coord}
\end{figure}

As seen in  Fig.~\ref{fig:coord}, such kinematic diversity is often a deliberate consequence of mechanical design rather than an arbitrary modeling choice. Humanoid legs must generate large torques within tightly constrained volumes while keeping distal mass and inertia low. Designers therefore relocate actuators away
from joint centers, employ linkage or parallel transmissions to obtain favorable leverage and distribute loads across actuators, and incline joint axes to improve workspace or locomotion-specific force transmission ~\cite{esser2021rh5,lutz2025actuation,fruend2022oblique}.
These choices can improve torque density and dynamic performance, but they also produce oblique, coupled, or configuration-dependent joint coordinates instead of a uniform body-aligned orthogonal arrangement.

In this study, we focus on the hypothesis that part of this embodiment dependence originates from the coordinates in which policy actions are expressed. Most locomotion policies output desired angles in the robot's native joint space~\cite{hwangbo2019agile,rudin2022minutes,xie2020cassie,radosavovic2024humanoid}. Consider a body-aligned hip-roll motion. If the hip axes are aligned with the pelvis frame, the motion may correspond mainly to one native joint. With oblique axes, the same motion requires coordinated changes of several native joints, even when they are independently actuated. Such axes may be introduced for packaging, range-of-motion, and mechanical-design considerations~\cite{fruend2022oblique}. The same task-level motion can therefore acquire a different numerical representation on each robot.

This issue is distinct from the conventional joint-to-actuator mapping. Four-bar linkages, parallel ankles, and other closed-chain transmissions are commonly handled by converting generalized joint targets into actuator commands~\cite{esser2021rh5,lutz2025actuation,sovukluk2023kangaroo}. These models describe how a robot-native joint motion is physically realized, but do not make the native coordinates themselves consistent across robots. Our method adds a virtual-to-native coordinate layer upstream of, rather than replacing, the existing transmission layer.

A sufficiently expressive policy can learn this relation implicitly, but then known embodiment-specific kinematics becomes embedded in the controller. A fixed orthogonal change of basis does not reduce the coverage of isotropic exploration. Practical locomotion pipelines, however, are not fully coordinate invariant: they commonly use factorized action distributions and per-joint scaling, clipping, gains, limits, and regularization~\cite{xie2020cassie,gu2024humanoidgym, schulman2017ppo}. Changing the native coordinates thus changes the correlations, relative scales, saturation margins, and penalties associated with an equivalent body-level motion.

Prior work has shown that action representation affects locomotion learning. Reference-relative joint targets provide useful structure~\cite{peng2018deepmimic}, task-space actions can improve learning efficiency~\cite{duan2021taskspace}, and torque actions exhibit different tuning and transfer characteristics from position targets~\cite{kim2023torque}. Cross-embodiment methods instead seek generalization through graph-structured or modular policies~\cite{wang2018nervenet,huang2020shared}, morphology-conditioned architectures~\cite{gupta2022metamorph,trabucco2022anymorph,xiong2023universal}, and multi-embodiment training~\cite{bohlinger2025multiembodiment}. These approaches are complementary to ours: we standardize the physical meaning of the joint-level action before robot-specific coordinate and transmission mappings are applied.

The contributions are threefold: (i) we identify native-coordinate inconsistency---distinct from the joint-to-actuator transmission, which existing models already handle---as a source of embodiment dependence, and address it with an analytic coordinate layer derived from the kinematic model that maps actions, joint-velocity observations, and joint-space gains
between a shared body-aligned space and each robot's native space; (ii) we show that where the map is inserted during training matters: applying it post-hoc avoids an instability that arises when it is composed in the training loop; and (iii) we evaluate the interface on three humanoids and on Unitree G1 hardware, where the benefit appears first in gait geometry rather than in aggregate task error.

\section{Background and Problem Formulation}
\label{sec:background}

\subsection{Native-Joint Policy Interface}

Most reinforcement-learning locomotion controllers formulate the policy action in the generalized joint coordinates provided by the robot model ~\cite{hwangbo2019agile,rudin2022minutes,xie2020cassie,
radosavovic2024humanoid}. Let $\mathbf q\in\mathcal Q^N$ and $\dot{\mathbf q}$ denote the native joint
configuration and velocity of a robot. Given an observation $\mathbf{o}\in\mathcal{O}$, the policy produces a desired joint configuration,
\begin{equation}
    \mathbf q^{\mathrm{des}}
    \sim
    \pi_{\theta}
    \left(
        \,\cdot\mid\mathbf{o}
    \right),
    \qquad
    \mathbf q^{\mathrm{des}}\in\mathcal Q^N .
    \label{eq:native_policy}
\end{equation}
The policy output is expressed as a desired joint angle; any normalization, output scaling, or range restriction is included in the policy parameterization. The superscript $N$ is used only for the configuration-space symbol $Q$ to distinguish the native spaces, and is omitted from other symbols for notational simplicity in this paper.

For directly actuated joints, the target is commonly tracked by a joint-space proportional--derivative (PD) controller,
\begin{equation}
    \boldsymbol{\tau}
    =
    \mathbf{K}_{p}
    \left(
        \mathbf q^{\mathrm{des}}-\mathbf{q}
    \right)
    -
    \mathbf{K}_{d}\dot{\mathbf{q}},
    \label{eq:native_pd}
\end{equation}
where $\mathbf{K}_p$ and $\mathbf{K}_d$ are control gains.
For robots with nontrivial transmissions, the desired generalized-joint configuration is first converted into an actuator control input $\mathbf{u}$,
\begin{equation}
    \mathbf u
    =
    \Psi
    \left(
        \mathbf x,
        \mathbf q^{\mathrm{des}}
    \right),
    \label{eq:joint_to_actuator}
\end{equation}
where $\mathbf x$ contains the states required by the transmission and low-level controller. The map $\Psi$ may be direct for a serial joint or nonlinear and configuration dependent for a four-bar linkage, parallel ankle, or another closed-chain mechanism ~\cite{esser2021rh5,lutz2025actuation, sovukluk2023kangaroo}.
In either case, the policy action remains defined in the generalized joint coordinates chosen for that robot.

This coordinate choice also shapes the learning problem: diagonal Gaussian action distributions~\cite{xie2020cassie}, action scales and bounds, servo gains, joint limits, and regularization are all defined per native joint, and none of these components is invariant to a change of joint coordinates.

\subsection{Problem Formulation}
\label{sec:problem}

Consider robot platforms $r$ and $s$ with native joint spaces
$\mathcal Q^{N}_r$ and $\mathcal Q^{N}_s$. Even when the robots have the same
joint dimensions, their native coordinates may represent different
rotation axes and axis orderings, as illustrated in
Fig.~\ref{fig:coord}. Let
$\Delta\mathcal M_r(\mathbf q_r,\Delta\mathbf q)$ denote the
body-relative link motion induced by a native-coordinate perturbation
$\Delta\mathbf q$. In general,
\begin{equation}
    \Delta\mathcal M_r(\mathbf q_r,\Delta\mathbf q)
    \neq
    \Delta\mathcal M_s(\mathbf q_s,\Delta\mathbf q).
    \label{eq:native_action_inconsistency}
\end{equation}
Thus, the same policy output can have different physical effects across
robots. A policy acting directly in native coordinates must therefore
learn not only the locomotion task but also the robot-specific
coordinate combinations required to produce task-relevant link
motions. Moreover, per-joint action distributions, output bounds,
control gains, and regularization make the resulting learning problem
dependent on the native coordinate convention.

The effects of joint geometry and native coordinates are not always visible in aggregate task metrics. With the same reward formulation, policies may achieve similar velocity tracking while producing asymmetric steps, irregular foot trajectories, or unnatural leg motions. Because such artifacts are poorly captured by standard scalar metrics, engineers often add and tune robot-specific reward terms through repeated inspection of simulation or hardware rollouts, limiting the practical reuse of a common training recipe.

Our objective is not identical whole-body motions, which also depend on geometry and dynamics, but a common interface in which corresponding policy dimensions represent the same body-aligned rotations with locally consistent physical effects, while preserving the action dimension, joint-target structure, and each robot's existing control stack.

\section{Proposed Method}
\label{sec:method}

\subsection{Architecture of Body-Aligned Virtual Joint Interface}
\label{sec:virtual_coordinates}

To address the coordinate inconsistency formulated in Sec.~\ref{sec:problem}, we introduce a common body-aligned virtual joint space and a robot-specific coordinate interface. For each robot platform
$r$, the policy outputs a desired virtual joint configuration, which is mapped to native joint targets before being passed to the robot's existing actuation interface.
Importantly, the virtual coordinates are not modeled as additional DoFs
in simulation. Because the virtual--native relation is determined by the robot kinematics
and branch convention, it is realized as a deterministic interface rather
than as extra simulated degrees of freedom, which preserves the original
dynamics and constraints and allows the same mapping to be used on
hardware.

\begin{figure}[t]
    \centering
    \includegraphics[width=0.95\linewidth]{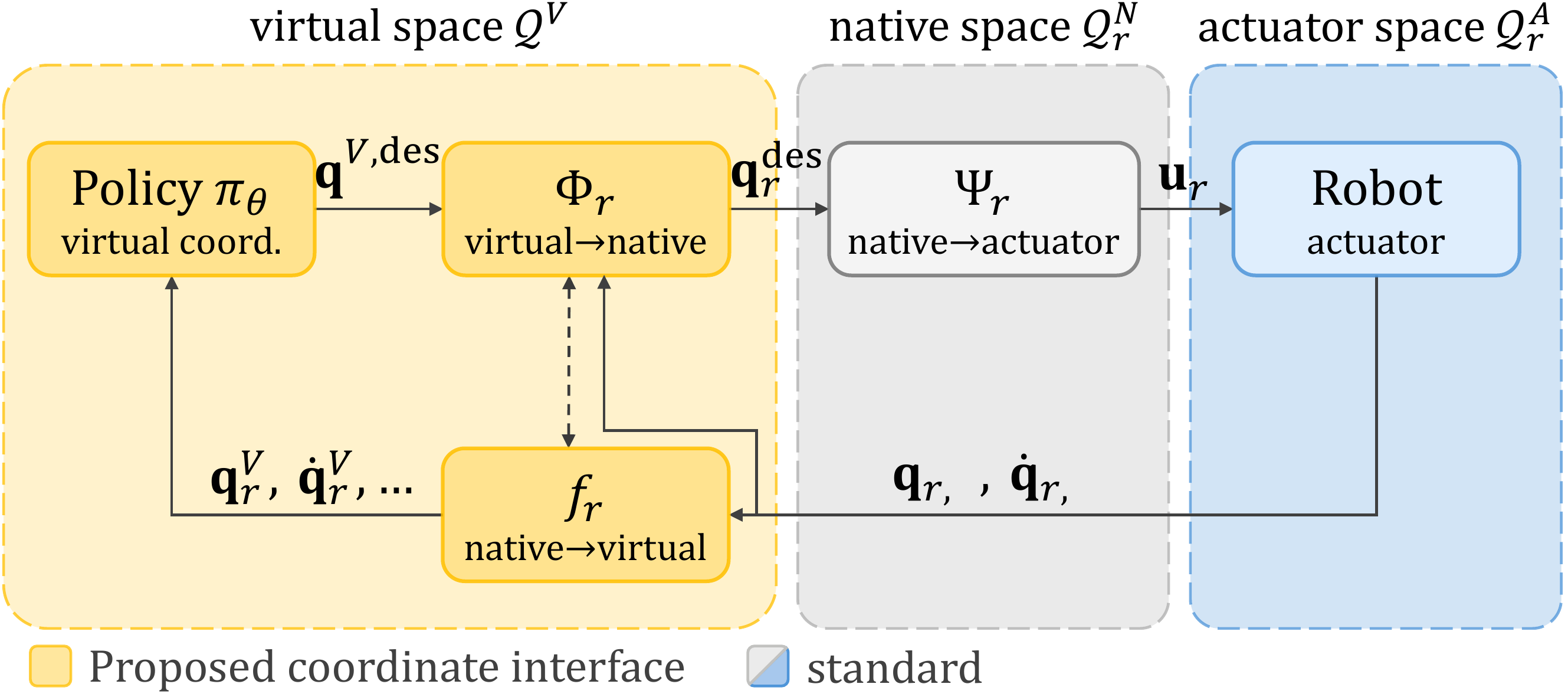}
    \caption{Overview of the proposed body-aligned virtual joint interface. The proposed modules map policy actions and joint-state observations between the shared virtual space and each robot's native space, while the existing native-to-actuator control stack remains unchanged.}
    \label{fig:framework}
\end{figure}

\subsubsection{Canonical humanoid-leg convention}
Let $\mathcal F_B$ denote the robot body frame, whose $x$-, $y$-, and $z$-axes point forward, leftward, and upward, respectively. For the common humanoid-leg topology, we define a canonical $3$--$1$--$2$ coordinate structure consisting of a 3-DoF hip, a 1-DoF knee, and a 2-DoF ankle:
\begin{equation}
    \mathbf q_{\mathrm{leg}}^{V}
    =
    \begin{bmatrix}
        \psi_{\mathrm{hip}} &
        \phi_{\mathrm{hip}} &
        \theta_{\mathrm{hip}} &
        \theta_{\mathrm{knee}} &
        \phi_{\mathrm{ankle}} &
        \theta_{\mathrm{ankle}}
    \end{bmatrix}^{\top}.
    \label{eq:canonical_leg_coordinates}
\end{equation}
Here, $(\psi,\phi,\theta)$ denote yaw, roll, and pitch about the body-aligned $z$-, $x$-, and $y$-axes. The corresponding orientation conventions are
\begin{equation}
\begin{aligned}
    \mathbf R_{\mathrm{hip}}^{V}
    &=
    \mathbf R_z(\psi_{\mathrm{hip}})
    \mathbf R_x(\phi_{\mathrm{hip}})
    \mathbf R_y(\theta_{\mathrm{hip}}),\\
    \mathbf R_{\mathrm{ankle}}^{V}
    &=
    \mathbf R_x(\phi_{\mathrm{ankle}})
    \mathbf R_y(\theta_{\mathrm{ankle}}).
\end{aligned}
\label{eq:canonical_group_rotations}
\end{equation}
Ankle yaw is not independently parameterized. For the knee, the nominal standing configuration is defined as zero, and the coordinate sign is chosen such that positive rotation corresponds to knee flexion.

\subsubsection{Link-based virtual-frame construction}
For each multi-DoF group $g\in\{\mathrm{hip},\mathrm{ankle}\}$, let $P_g$ and $D_g$ denote the
proximal and distal links bounding its native joint chain. These are the pelvis and thigh for the hip, and the shank and foot for the ankle. A virtual reference frame is rigidly attached to each link. At the nominal configuration $\mathbf q_r^0$, both frames are aligned with $\mathcal F_B$; thereafter, they move with their respective links. Their relative orientation therefore describes thigh motion relative to the pelvis or foot motion relative to the shank in a common body-aligned
convention.

For a URDF-based implementation, let ${}^{B}\mathbf R_{r,X}(\mathbf q_r)$ denote the orientation of link frame $X$ obtained by forward kinematics. The constant calibration offsets are
\begin{equation}
    \mathbf C_{r,X}
    =
    \left(
        {}^{B}\mathbf R_{r,X}(\mathbf q_r^0)
    \right)^{\top},
    \qquad
    X\in\{P_g,D_g\},
    \label{eq:virtual_frame_calibration}
\end{equation}
and the calibrated relative orientation is
\begin{equation}
\begin{aligned}
    \overline{\mathbf R}_{r,g}(\mathbf q_{r,g})
    &=
    \left(
        {}^{B}\mathbf R_{r,P_g}(\mathbf q_r)
        \mathbf C_{r,P_g}
    \right)^{\top} \\
    &\quad
    {}^{B}\mathbf R_{r,D_g}(\mathbf q_r)
    \mathbf C_{r,D_g}.
\end{aligned}
\label{eq:calibrated_relative_orientation}
\end{equation}
By construction, $\overline{\mathbf R}_{r,g}(\mathbf q_{r,g}^{0})=\mathbf I$. Although the forward kinematics may be evaluated using the complete configuration $\mathbf q_r$, the relative orientation depends only on the joints between $P_g$ and $D_g$. Because only orientation is used,
the link-frame origins do not affect the virtual coordinates. The same construction applies to any kinematic model that provides the required link transforms.

\subsubsection{Native--virtual coordinate encoder}
For compactness, let $\mathcal G=\{\mathrm{hip},\mathrm{knee},\mathrm{ankle}\}$, with virtual dimensions $3$, $1$, and $2$, respectively. Let $\mathcal Q_{r,g}^{N}$ and $\mathcal Q_g^V$ denote the native and virtual configuration spaces of group $g$. The robot kinematics define an encoding map
\begin{equation}
    f_{r,g}:
    \mathcal Q_{r,g}^N
    \rightarrow
    \mathcal Q_g^V,
    \qquad
    f_{r,g}(\mathbf q_{r,g}^{0})=\mathbf 0.
    \label{eq:virtual_encoder}
\end{equation}
For the hip and ankle, $f_{r,g}$ extracts the coordinates of $\overline{\mathbf R}_{r,g}$ according to Eq.~\eqref{eq:canonical_group_rotations}; for the knee, it applies only the calibrated zero offset and sign. Thus, corresponding virtual coordinates have the same anatomical interpretation across robots, without requiring identical native joint layouts.

\subsubsection{Insertion point of the interface}
The interface can be introduced at two different points of the pipeline, and we distinguish them explicitly because they lead to different optimization behavior. In the \emph{in-loop} form, $f_r$ and $\Phi_r$ are evaluated at every control step of training on the unmodified native model, so the policy interacts with the native kinematics through the coordinate layer from the first iteration. In the \emph{post-hoc} form, which is the configuration proposed in this work, training is instead carried out on a body-aligned model of the same robot---the kinematic model with each leg joint axis replaced by the corresponding canonical
axis of Eq.~(5), leaving link masses, inertias, and joint limits unchanged---so that native and virtual coordinates coincide during training and no coordinate map is evaluated in the loop. The maps $f_r$ and $\Phi_r$ are then applied at deployment to run the resulting policy
on the native robot. Both forms share the virtual action semantics and deployment stack; the
post-hoc form instead trains on an idealized model and absorbs the resulting mismatch as a deployment-time disturbance, which Sec.~IV-B-3 shows is benign relative to the instability of the in-loop form.

\subsection{Orientation-Based Native Mapping}
\label{sec:fixed_mapping}
This subsection considers the nontrivial hip and ankle mappings; the calibrated knee map is one-dimensional and affine. Given a desired virtual configuration $\mathbf q_g^{V,\mathrm{des}}$, the corresponding native target is defined by orientation matching:
\begin{equation}
    \overline{\mathbf R}_{r,g}
    \left(
        \mathbf q_{r,g}^{\mathrm{des}}
    \right)
    =
    \mathbf R_g^{V}
    \left(
        \mathbf q_g^{V,\mathrm{des}}
    \right).
    \label{eq:orientation_matching}
\end{equation}
When the native joint axes span the group's rotational DoF, so that the matched orientation maps to a unique joint-coordinate vector, the native target is obtained directly as
\begin{equation}
    \mathbf q_{r,g}^{\mathrm{des}}
    =
    \operatorname{coord}_{r,g}
    \left[
        \mathbf R_g^{V}
        \left(
            \mathbf q_g^{V,\mathrm{des}}
        \right)
    \right]
    \equiv
    \Phi_{r,g}
    \left(
        \mathbf q_g^{V,\mathrm{des}}
    \right),
    \label{eq:orientation_mapping}
\end{equation}
where $\operatorname{coord}_{r,g}(\cdot)$ extracts the native joint coordinates according to the robot-specific axis arrangement and joint ordering. When multiple solutions exist, the branch continuous with the nominal configuration is selected. The relation between finite native and virtual angles is generally nonlinear; hence, no constant linear transformation between their joint-angle vectors is assumed.

\subsection{Differential Mapping by Full-Coordinate Jacobian}
\label{sec:configuration_mapping}
The group-wise encoders are assembled into a single robot-level coordinate interface:
\begin{equation}
    \mathbf q^{V}
    =
    f_r(\mathbf q_r)
    =
    \operatorname{col}_{g\in\mathcal G}
    \left[
        f_{r,g}(\mathbf q_{r,g})
    \right].
    \label{eq:full_coordinate_map}
\end{equation}
Groups that already follow the virtual convention are represented by identity components of $f_r$. The corresponding coordinate Jacobian is
\begin{equation}
    \mathbf J_r(\mathbf q_r)
    =
    \frac{\partial f_r}{\partial\mathbf q_r},
    \qquad
    \dot{\mathbf q}^{V}
    =
    \mathbf J_r(\mathbf q_r)\dot{\mathbf q}_r.
    \label{eq:full_coordinate_jacobian}
\end{equation}
Because each virtual group depends only on its corresponding native group,

\begin{equation}
    \mathbf J_r(\mathbf q_r)
    =
    \operatorname{blkdiag}_{g\in\mathcal G}
    \left[
        \mathbf J_{r,g}(\mathbf q_{r,g})
    \right].
    \label{eq:block_coordinate_jacobian}
\end{equation}
For a single leg:
\begin{equation}
    \mathbf J_r
    =
    \operatorname{blkdiag}
    \left(
        \mathbf J_{r,\mathrm{hip}},
        s_{r,\mathrm{knee}},
        \mathbf J_{r,\mathrm{ankle}}
    \right),
    \label{eq:leg_coordinate_jacobian}
\end{equation}
where $\mathbf J_{r,\mathrm{hip}}\in\mathbb R^{3\times3}$, $\mathbf J_{r,\mathrm{ankle}}\in\mathbb R^{2\times2}$, and $s_{r,\mathrm{knee}}\in\{-1,1\}$ accounts for the calibrated direction.

For each multi-DoF group, let $\mathbf E_g^{V}(\mathbf q_g^{V})$ and $\mathbf E_{r,g}(\mathbf q_{r,g})$ map virtual and native joint rates to the same angular velocity expressed in the proximal-link frame:
\begin{equation}
    \boldsymbol{\omega}_g^{P}
    =
    \mathbf E_g^{V}(\mathbf q_g^{V})
    \dot{\mathbf q}_g^{V}
    =
    \mathbf E_{r,g}(\mathbf q_{r,g})
    \dot{\mathbf q}_{r,g}.
    \label{eq:orientation_jacobians}
\end{equation}
When $\mathbf E_g^{V}$ has full column rank,
\begin{equation}
    \mathbf J_{r,g}(\mathbf q_{r,g})
    =
    \left[
        \mathbf E_g^{V}
        \left(
            f_{r,g}(\mathbf q_{r,g})
        \right)
    \right]^{\dagger}
    \mathbf E_{r,g}(\mathbf q_{r,g}),
    \label{eq:group_coordinate_jacobian}
\end{equation}
where $(\cdot)^\dagger$ denotes the Moore--Penrose pseudoinverse. For a 3-DoF rotational group away from a representation singularity, it reduces to the ordinary inverse. For the 2-DoF ankle, the relation is defined on its admissible rotational subspace.

For the yaw--roll--pitch hip convention,
\begin{equation}
    \mathbf E_{\mathrm{hip}}^{V}
    =
    \left[
        \mathbf e_z,\;
        \mathbf R_z(\psi)\mathbf e_x,\;
        \mathbf R_z(\psi)\mathbf R_x(\phi)\mathbf e_y
    \right],
    \label{eq:virtual_orientation_jacobian}
\end{equation}
whose determinant is $\det\mathbf{E}^{V}_{\mathrm{hip}}=\cos\phi_{\mathrm{hip}}$.
The representation becomes singular only at $\phi=\pm 90^\circ$, where the
yaw and pitch axes are collinear.
Choosing roll as the middle rotation places this singularity outside the hip-abduction range encountered during normal locomotion. Robot-specific mechanism singularities may additionally enter through $\mathbf E_{r,g}$.

The coordinate Jacobian can also be used to map a common virtual-space stiffness to the native coordinates. Around the nominal configuration,
\begin{equation}
    \mathbf K_r^{\mathrm{full}}
    =
    \mathbf J_r^\top(\mathbf q_r^0)
    \mathbf K^V
    \mathbf J_r(\mathbf q_r^0).
    \label{eq:gain_mapping}
\end{equation}
For independent joint gains, we use its diagonal approximation $\mathbf K_r=\operatorname{Diag}(\operatorname{diag}(\mathbf K_r^{\mathrm{full}}))$. The same mapping applies to damping gains.

The full Jacobian is also used to transform joint-velocity observations. When an explicit decoder is unavailable for a particular mechanism, it may support a damped local numerical inverse. The standard pseudoinverse is used in regular configurations, with damping  introduced only near rank-deficient configurations.

\subsection{Policy, Observation, and Actuation Integration}
\label{sec:policy_integration}
The group-wise decoders are assembled into the full map $\Phi_r$, with identity mappings for unaffected joints. The complete action path is
\begin{equation}
\begin{aligned}
    \mathbf q_r^{\mathrm{des}}
    &=
    \Phi_r
    \left(
        \mathbf q^{V,\mathrm{des}};
        \mathbf q_r
    \right),\\
    \mathbf u_r
    &=
    \Psi_r
    \left(
        \mathbf x_r,
        \mathbf q_r^{\mathrm{des}}
    \right),
\end{aligned}
\label{eq:complete_pipeline}
\end{equation}
where $\Psi_r$ is the robot's existing native-joint-to-actuator interface and $\mathbf x_r$ contains any state required by the existing controller. The current configuration is included in $\Phi_r$ only when needed for branch selection or a configuration-dependent inverse. The transmission model, low-level controller, safety checks, and physical joint limits remain unchanged.
Closed-chain mechanisms such as four-bar linkages and parallel ankles, present on the evaluated platforms, are handled entirely within $\Psi_r$ as in the native pipeline; the proposed layer acts upstream on the joint-coordinate semantics and is independent of how those native targets are physically realized.

Measured joint states are encoded using the same convention:
\begin{equation}
    \mathbf q_{r,g}^{V}
    =
    f_{r,g}(\mathbf q_{r,g}),
    \qquad
    \dot{\mathbf q}_{r,g}^{V}
    =
    \mathbf J_{r,g}(\mathbf q_{r,g})
    \dot{\mathbf q}_{r,g}.
    \label{eq:virtual_observation}
\end{equation}
Joint-position and joint-velocity observations, action bounds, and action-related regularization are expressed in virtual coordinates. Native joint limits and actuator saturation are enforced after the mapping in their physical spaces. The proposed and native-coordinate controllers therefore retain the same policy architecture and action dimension; only the joint-coordinate interface is changed.

\section{Simulation and Experimental Evaluation}
\label{sec:experiments}

\begin{figure}[t]
    \centering
    \includegraphics[width=0.7\linewidth]{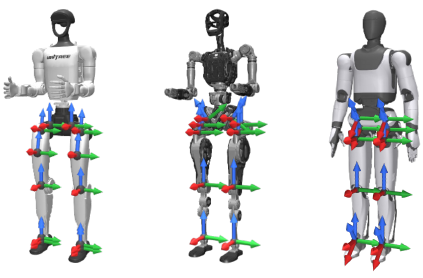}
    \caption{Humanoid platforms used in simulation—Unitree G1, PND Adam, and a custom-built platform—with their leg joint-axis configurations.}
    \label{fig:robots}
\end{figure}

We evaluate whether the proposed body-aligned interface enables a single locomotion recipe to generalize across humanoids with different native joint layouts. We compare it with native- and task-space baselines in simulation and assess transfer to hardware.

\subsection{Platforms and Experimental Setup}

\label{sec:setup}

\subsubsection{Platforms and Training Setup}
We evaluate the proposed interface on three humanoid platforms with distinct leg kinematics: the Unitree G1, the PND Adam, and a custom-built platform (Fig.~\ref{fig:robots}). Each platform is used in its $12$-DoF bipedal lower-body configuration, with six joints per leg following the canonical $3$--$1$--$2$ topology. For every platform, the body-aligned virtual coordinates follow the canonical body convention. The per-robot encoder $f_r$, decoder $\Phi_r$ (Sec.~\ref{sec:virtual_coordinates}), and Jacobian matrix are derived analytically from the URDF model.

All configurations share a single training recipe, implemented as a manager-based velocity-tracking environment with GPU-parallel simulation in Isaac Lab~\cite{rudin2022minutes}. Unless otherwise specified, the network architecture, PPO hyperparameters, observations, commands, rewards, curriculum, events, and simulation settings follow the official Unitree RL Lab G1-29DoF velocity-tracking configuration~\cite{unitree_rl_lab}. To construct the 12-DoF leg-only controller used in this study, the arm
and waist joints and their associated reward terms are removed, and the
upper-body joints are fixed at their nominal initial configuration during training. A gait-phase clock is added only to the critic observation, while all remaining settings are kept unchanged.
During sim-to-sim and hardware deployment, the upper-body joints are controlled independently to maintain the same initial posture. Across all three robot platforms and coordinate configurations, only the joint-coordinate interface used to express the policy actions and joint-state observations is varied, thereby isolating the effect of the proposed coordinate representation.

\subsubsection{Baselines and Evaluation Protocol}
Against this fixed recipe we compare four joint-coordinate interfaces. (i) \emph{Native joint space} is the
conventional baseline, in which the policy acts and observes directly in each robot's native joint coordinates. (ii) \emph{Task space} follows Duan et al.~\cite{duan2021taskspace}: the policy outputs foot-position residuals and direct hip-yaw and foot-pitch targets, which are converted to joint-level commands by a model-based task-space controller.
The remaining two interfaces both use the body-aligned virtual joint space
and differ only in where the coordinate map is inserted (Sec.~III-A):
(iii) \emph{Virtual joint space (in-loop)}, in which the analytic coordinate map is applied on the unmodified native model throughout training, and (iv) \emph{Virtual joint space (post-hoc)}, in which the policy is trained on a body-aligned model and the coordinate map is applied to the native robot afterward. Note that \emph{Virtual joint space (post-hoc)} is the proposed method in this study.

All reported metrics are measured under sim-to-sim transfer: each trained policy is evaluated closed-loop in MuJoCo~\cite{todorov2012mujoco}, whose physics and contact model differ from the Isaac Lab training environment. The full deployment stack runs unchanged: $\Phi_r$ decodes virtual targets to native targets, which each robot's PD controller and transmission convert to actuator commands. Results are averaged over diverse velocity commands and $n=3$ seeds.

\subsection{Learning Performance Across Joint Layouts}

\begin{table}[t]
\centering
\caption{Sim-to-sim velocity tracking on the constant velocity command ($n=3$ seeds, mean$\pm$std). Velocity error and lateral drift are expressed in the base heading frame.}
\label{tab:tracking}
\footnotesize
\setlength{\tabcolsep}{4pt}
\begin{tabular}{@{}llcc@{}}
\toprule
Robot & Interface & Velocity Error (m/s) & Lateral Drift (m) \\
\midrule
\multirow{4}{*}{Unitree G1}
 & Native             & \textbf{0.042\,$\pm$\,0.001} & 0.086\,$\pm$\,0.001 \\
 & Task space         & 0.048\,$\pm$\,0.002 & 0.129\,$\pm$\,0.003 \\
 & Virtual (in-loop)  & \textbf{0.042\,$\pm$\,0.007} & \textbf{0.084\,$\pm$\,0.001} \\
 & Virtual (post-hoc) & 0.053\,$\pm$\,0.018 & 0.089\,$\pm$\,0.003 \\
\midrule
\multirow{4}{*}{PND Adam}
 & Native             & 0.106\,$\pm$\,0.018 & 0.094\,$\pm$\,0.007 \\
 & Task space         & 0.207\,$\pm$\,0.016 & 0.082\,$\pm$\,0.010 \\
 & Virtual (in-loop)  & 0.143\,$\pm$\,0.034 & 0.113\,$\pm$\,0.013 \\
 & Virtual (post-hoc) & \textbf{0.085\,$\pm$\,0.018} & \textbf{0.075\,$\pm$\,0.006} \\
\midrule
\multirow{4}{*}{\makecell{Custom-built\\Platform}}
 & Native             & 0.032\,$\pm$\,0.005 & 0.042\,$\pm$\,0.002 \\
 & Task space         & 0.477\,$\pm$\,0.333 & 0.120\,$\pm$\,0.070 \\
 & Virtual (in-loop)  & 0.033\,$\pm$\,0.005 & 0.040\,$\pm$\,0.003 \\
 & Virtual (post-hoc) & \textbf{0.029\,$\pm$\,0.002} & \textbf{0.033\,$\pm$\,0.005} \\
\bottomrule
\end{tabular}
\end{table}

\subsubsection{Velocity Tracking and Lateral Drift}
Under the single unmodified recipe, all four interfaces learn to track the commanded velocities on all three robots.
We evaluate tracking under sim-to-sim transfer using a predefined set of 25 constant velocity-command combinations spanning $v_x\in[-0.5,1.0]\,\mathrm{m/s}$, $v_y\in[-0.3,0.3]\,\mathrm{m/s}$, and $\omega_z\in[-0.2,0.2]\,\mathrm{rad/s}$. The results are averaged over all 25~commands, reporting the velocity-tracking error.
Additional sim-to-sim tests evaluate lateral drift, defined as the magnitude of the lateral displacement accumulated 20\,s rollout in the initial base-heading frame under a straight command ($v_x=0.5$, $v_y=\omega_z=0$).

As can be seen in Table~\ref{tab:tracking}, on the two obliquely-hipped platforms, Adam and the custom robot, the proposed interface attains both the lowest tracking error and the smallest lateral drift of all interfaces;
on G1, whose native hip carries only a small ($\sim 10^\circ$) mounting offset and is therefore already close to the body convention, the interfaces track comparably, with the differences falling within the seed-to-seed spread and leaving little for the coordinate change to correct.
The task-space baseline instead degrades on Adam and fails on the custom platform, where it barely advances under the command ($0.48\,$m/s error). The takeaway is not that the proposed interface dominates everywhere, but that a single unchanged recipe attains accurate, low-drift tracking across all three embodiments, whereas each baseline degrades on at least one platform.

\subsubsection{Gait-Pattern Consistency}

Accurate velocity tracking does not necessarily imply a well-aligned gait motion. Policies can satisfy the commanded velocity while producing unintended and visually unnatural gait patterns, such as persistent toe-in or toe-out, particularly with oblique native hip axes. We therefore evaluate foot alignment using the foot-progression angle (FPA), defined as the foot heading relative to the base heading during mid-stance:
\begin{equation}
\begin{aligned}
    \mathrm{FPA} &= \operatorname{atan2}\!\big(r_{21}^{f},\,r_{11}^{f}\big)-\psi_b, \\[2pt]
    \overline{|\mathrm{FPA}|} &= \frac{1}{2}\sum_{f\in\{L,R\}}\big\langle|\mathrm{FPA}_f|\big\rangle,
\end{aligned}
\label{eq:fpa}
\end{equation}
where $\mathbf R_f=[r_{ij}^{f}]$ is the foot rotation matrix,
$\psi_b$ is the base heading, and $\langle\cdot\rangle$ denotes averaging over mid-stance samples and gait cycles. A value close to $0^\circ$ indicates that the feet remain aligned with the body's forward direction.

\begin{figure}[t]
\centering
\includegraphics[width=\linewidth]{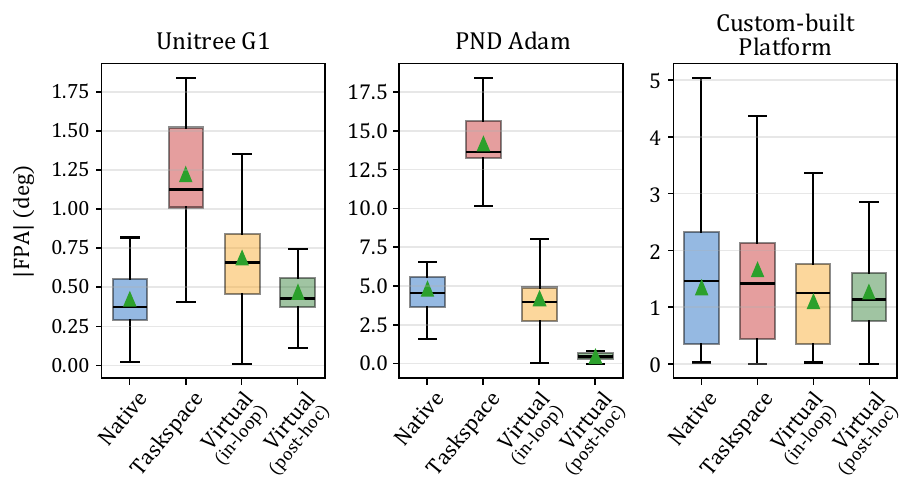}
\caption{Foot-progression angle (toe-out) by platform and coordinate interface (all seeds, both feet).
}
\label{fig:toeout}
\end{figure}

\begin{figure}[!t]
\centering
\includegraphics[width=0.95\linewidth]{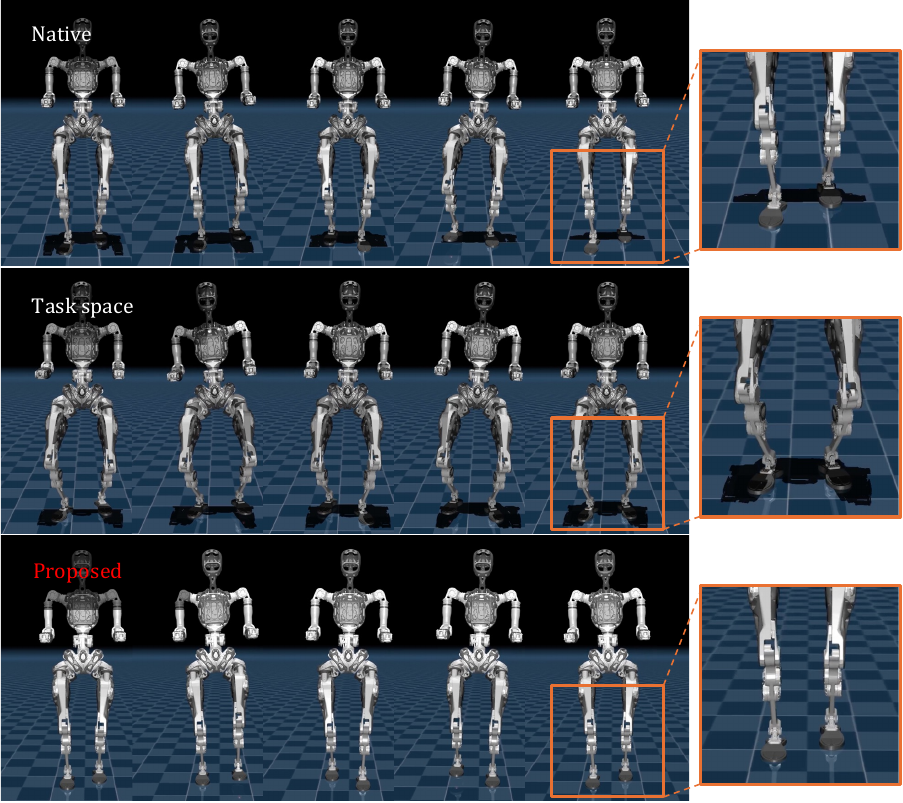}
\caption{Swing-leg motion on the PND Adam in simulation under the native, proposed, and task-space interfaces.
The proposed interface keeps the feet aligned with the direction of travel.}
\label{fig:adam_swing}
\end{figure}

The proposed interface keeps the feet essentially aligned with the body on every
platform ($\overline{|\mathrm{FPA}|}\le1.3^\circ$), whereas the baselines exhibit
a strongly \emph{embodiment-dependent} toe-out (Fig.~\ref{fig:toeout}). The effect
is largest on the Adam platform. Its hip roll and yaw axes are pelvis-aligned at
the nominal pose, but the \emph{proximal} hip-pitch axis is inclined by $35^\circ$
about the fore--aft ($x$) axis, with mirrored signs on the two legs. Because the
tilted axis is both the most proximal joint of the chain and the one carrying the
largest excursion in walking, a native hip-pitch command of $\theta$ injects
$\sin(35^\circ)\,\theta\approx0.57\theta$ of body yaw of opposite sign on each
leg---at a typical $23^\circ$ swing, $13^\circ$ of thigh yaw---producing an
outward splay that native coordinates can cancel only by learning the axis mixing
implicitly. The native and task-space interfaces produce mean foot-progression
angles of $4.8^\circ$ and $14.2^\circ$, respectively, compared with $0.5^\circ$
for the proposed interface, as also seen qualitatively in
Fig.~\ref{fig:adam_swing}. On G1 and the custom platform the mounting tilt is
instead about the lateral ($y$) axis and shared by both legs; it leaves the pitch
axis invariant and mixes only roll and yaw, whose excursions are small in steady
walking, so toe-out is small for all interfaces ($\le1.7^\circ$).

The proposed interface therefore does not reduce toe-out universally;
rather, it prevents the pronounced outward foot misalignment induced by
native coordinates on platforms with obliquely mounted hips.

Foot alignment alone does not capture temporal coordination or step-to-step repeatability. Let $t_k^f$ denote the $k$-th lift-off of foot $\{L,R\}$, and let $\tau_k^R$ be the first right-foot lift-off following $t_k^L$.
We measure gait-period variability using the coefficient of variation $\mathrm{CV}(T)$ and deviation from ideal left--right alternation as
\begin{equation}
e_\Phi = 100\left\langle\left|\frac{\tau^R_k-t^L_k}{\widetilde{T}}-\frac{1}{2}\right|\right\rangle_k ,
\end{equation}
where $\widetilde{T}$ is the median gait period.
Thus, $e_{\Phi}=0$ indicates exact alternation. All metrics are averaged over steady-state rollouts from three seeds.

The proposed interface provides more consistent inter-leg coordination than native coordinates across all three platforms. It reduces the phase error from $1.62\%$ to $0.23\%$ on G1 and from $1.64\%$ to $0.54\%$ on the custom platform, while reducing it only marginally on Adam (from $2.59\%$ to $2.36\%$). Thus, although the improvement on Adam is
modest, the proposed interface consistently matches or outperforms native coordinates without platform-specific tuning.
The other interfaces exhibit more pronounced embodiment-dependent behavior. On the custom platform, the task-space baseline reaches $\mathrm{CV}(T)=11.4\%$, whereas the proposed and native-coordinate interfaces remain below $0.6\%$. On Adam, one seed of the in-loop variant converges to a substantially shorter gait period ($0.668$~s versus approximately $0.81$~s for the other seeds), consistent with its unstable training behavior. The task-space baseline also converges to a near-extended-knee gait on G1 and the custom platform, indicating a further platform-dependent failure under the shared training recipe.

These gait-shape differences are not penalized by a velocity-tracking objective, yet in practice they are exactly what practitioners suppress through manual, often platform-specific, reward shaping. Because the proposed interface produces body-consistent gaits as an emergent property of the coordinate representation rather than of a tuned reward, it removes a recurring and time-consuming step from the locomotion-learning workflow---a benefit that, in our experience, is as valuable as the tracking results themselves.

\subsubsection{Training Robustness Across Embodiments}

We evaluate training robustness using episode-length stability rather than wall-clock time, since the runs used different GPU resources. The instability fraction is defined as the proportion of training iterations, counted after the mean episode length first reaches half of its maximum, in which it subsequently falls back below that level; the initial learning
transient is therefore excluded. We do not compare raw rewards because the coordinate interfaces alter the scales of action-dependent penalties.

\begin{figure}[t]
\centering
\includegraphics[width=\linewidth]{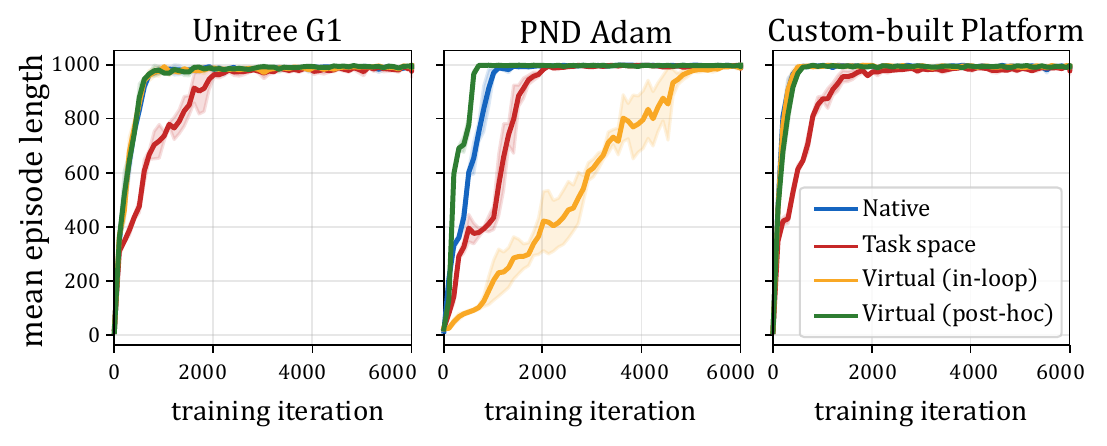}
\caption{Training stability across the three embodiments, measured by mean episode length over the first $6$k iterations (mean $\pm$ standard deviation over three seeds). The dotted line marks half of the maximum episode length.}
\label{fig:stability}
\end{figure}

As shown in Fig.~\ref{fig:stability}, the proposed interface trains consistently across all three embodiments, remaining in the unstable regime for at most $0.1\%$ of the training iterations. In contrast, the in-loop variant reaches an instability fraction of $15.3\%$ on Adam, while the task-space baseline remains unstable for $1.2$--$6.0\%$ of training across the three platforms. These results show that training on the body-aligned model and applying the coordinate map afterward avoids the embodiment-dependent optimization instability observed with the alternative interfaces.

\subsection{Hardware Experiments}

\begin{figure}[t]
\centering
\includegraphics[width=0.95\columnwidth]{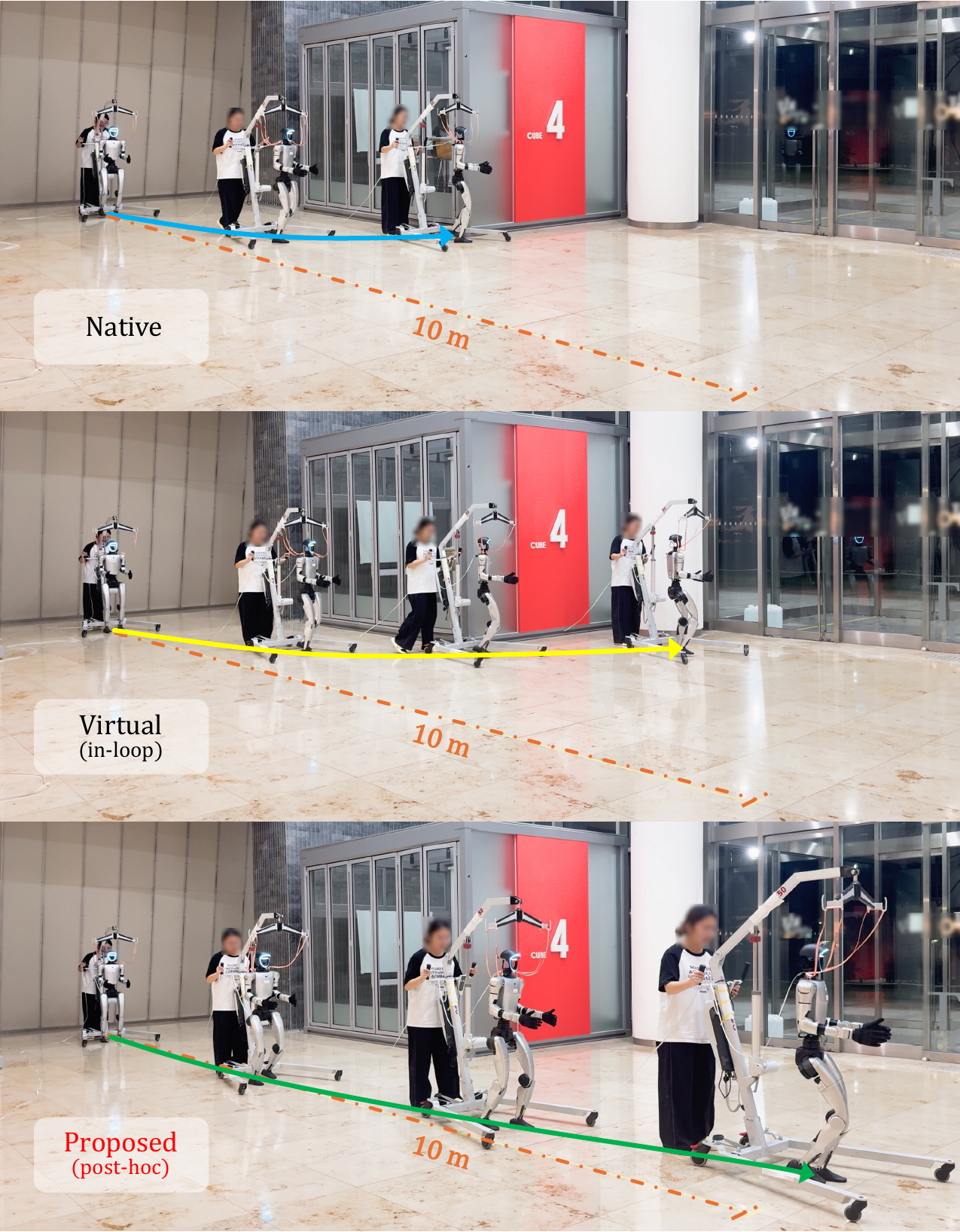}
\caption{
Hardware rollouts under a constant straight-walking command, composited from video frames of one representative rollout per interface. The dash-dotted line is a $10\,$m floor reference along the commanded heading; the solid curve traces the path walked.
}
\label{fig:hw_paths}
\end{figure}

\subsubsection{Deployment Feasibility and Path Regularity}

We deployed the native, in-loop, and proposed interfaces on the Unitree G1 without reward retuning, gain adjustment, or coordinate-specific modification. Each controller was commanded to walk straight at $v_x=0.5\,\mathrm{m/s}$ with $v_y=\omega_z=0$, over $6$ rollouts each.

All three interfaces produced stable walking, but differed substantially in their ability to maintain the commanded heading. The native and in-loop controllers drifted toward the edge of the test area and were stopped after $5.1\pm1.7\,\mathrm{m}$ and $6.9\pm3.0\,\mathrm{m}$, respectively. In contrast, every rollout of the proposed interface reached the end of the available walkway, traveling $13.8\pm1.1\,\mathrm{m}$; this distance was therefore limited by the test area rather than controller failure.

Because base position was not available from the deployment logs, heading drift was quantified using the absolute IMU yaw rate over an identical $5$-s steady-walking interval. The proposed interface reduced the yaw rate to $0.89\pm0.54^\circ/\mathrm{s}$, compared with
$5.31\pm1.90^\circ/\mathrm{s}$ for native coordinates and $4.08\pm1.02^\circ/\mathrm{s}$ for the in-loop interface. Although the interfaces exhibited similar aggregate tracking performance on G1 in simulation, the hardware results reveal a clear difference in their
ability to preserve the commanded heading under the sim-to-real gap.

The separation on hardware, absent from the aggregate sim-to-sim tracking of Table~\ref{tab:tracking}, admits a consistent reading. Heading drift is a signed, accumulating quantity: a small out-of-plane component of the swing foot injects a yaw impulse of the same sign at every step, so a bias that a magnitude-averaged velocity error cannot resolve integrates into meters of deviation over a $10\,\mathrm{m}$ walkway (Fig.~\ref{fig:hw_paths}). The separation is in fact already present in simulation, but confined to gait geometry---the inter-leg phase error on the same platform falls from $1.62\%$ to $0.23\%$---rather than to task error. What changes on hardware is the perturbation acting on the coordinate layer itself. Joint-axis calibration residuals, mounting tolerances, transmission compliance, and load-dependent tracking lag each insert a small quasi-static rotation between the commanded and the realized link orientation. Expressed in obliquely mounted native axes, such a discrepancy gives a purely sagittal intent a systematic abduction and yaw component, which the policy can only offset through the axis mixing it has implicitly learned from the nominal simulated geometry---precisely the assumption the real robot violates. The proposed interface instead supplies this mixing analytically through $f_r$ and $\Phi_r$, leaving the residual as an unmodeled disturbance
on a body-aligned action rather than a bias in a learned compensation. A controlled attribution would require hardware ablations beyond these rollouts.

\subsubsection{Onboard Computational Cost}
On the Jetson Orin NX, the virtual-coordinate encoder and decoder require approximately $0.2\,\mathrm{ms}$ per control step, corresponding to about $1\%$ of the $20\,\mathrm{ms}$ control period. Including the gain transformation of Eq.~\eqref{eq:gain_mapping} and policy evaluation, the complete control thread requires approximately $1.10\,\mathrm{ms}$ for
the proposed interface and remains compatible with the $50$-Hz control rate. No control deadline was missed during any hardware rollout.

\section{Conclusion}

We introduced a body-aligned virtual joint interface that maps shared policy actions and observations to each robot's native joint space while preserving the existing transmission and low-level control stack. Across three humanoids with different native leg-axis layouts, a single unmodified recipe produced accurate, body-aligned, and stably trained velocity-tracking policies, whereas each baseline degraded on at least one platform. On Unitree G1 hardware, where simulation reports the smallest difference between interfaces, it reduced heading drift by nearly a factor of six at negligible onboard cost. A shared body-aligned convention can therefore reduce coordinate-induced embodiment dependence without per-robot tuning, and the benefit is understated by aggregate task metrics in simulation.

This study considered only legged velocity-tracking locomotion, with three seeds per configuration and hardware validation on a single platform. Future work will extend the interface to the arms, waist, and torso and investigate whole-body motion imitation and retargeting, enabling both training recipes and body-aligned reference motions to be reused across heterogeneous humanoids.

\bibliographystyle{IEEEtran}
\bibliography{MyReference}

\end{document}